\documentclass[letterpaper, 10 pt, conference]{ieeeconf}  

\IEEEoverridecommandlockouts                              

\usepackage{graphics} 
\usepackage{epsfig} 
\usepackage{mathptmx} 
\usepackage{times} 
\usepackage{amsmath} 
\usepackage{amssymb}  
\usepackage{graphicx}
\usepackage{algorithm}
\usepackage{algorithmic}
\usepackage{bm}
\usepackage{wrapfig}
\usepackage{placeins}
\usepackage{multirow}
\usepackage{booktabs}
\usepackage{float}
\usepackage[dvipsnames]{xcolor}
\usepackage{hyperref}
\usepackage[normalem]{ulem}

\DeclareMathAlphabet{\mathcal}{OMS}{cmsy}{m}{n}

\renewcommand{\algorithmiccomment}[1]{\color{teal}{$\rhd$ #1}\color{black}} %

\usepackage{etoolbox}
\makeatletter
\patchcmd{\@makecaption}
  {\scshape}
  {}
  {}
  {}
\makeatother

\usepackage{setspace}

\newcommand{\jie}[1]{{\color{red}[Jie: #1]}}

\renewcommand{\baselinestretch}{0.97}

\title{\LARGE \bf
Multi-Objective Graph Heuristic Search for \\ Terrestrial Robot Design
}

\author{Jie Xu, Andrew Spielberg, Allan Zhao, Daniela Rus, Wojciech Matusik \\
\href{http://moghs.csail.mit.edu}{\color{magenta}http://moghs.csail.mit.edu}
\thanks{J. Xu, A. Spielberg, A. Zhao, D. Rus, W. Matusik are with the MIT Computer Science And Artificial Intelligence Laboratory (CSAIL), Massachusetts Institute of Technology, 77 Massachusetts Avenue,
Cambridge, Massachusetts, \url{
{jiex, aespielberg}@csail.mit.edu, azhao@mit.edu, {rus, wojciech}@csail.mit.edu}
}
}

\begin{document}

\maketitle
\thispagestyle{empty}
\pagestyle{empty}

\begin{abstract}
We present methods for co-designing rigid robots over control and morphology (including discrete topology) over multiple objectives.  Previous work has addressed problems in single-objective robot co-design or multi-objective control. However, the joint multi-objective co-design problem is extremely important for generating capable, versatile, algorithmically designed robots. In this work, we present Multi-Objective Graph Heuristic Search, which extends a single-objective graph heuristic search from previous work to enable a highly efficient multi-objective search in a combinatorial design topology space.  Core to this approach, we introduce a new universal, multi-objective heuristic function based on graph neural networks that is able to effectively leverage learned information between different task trade-offs. We demonstrate our approach on six combinations of seven terrestrial locomotion and design tasks, including one three-objective example. We compare the captured Pareto fronts across different methods and demonstrate that our multi-objective graph heuristic search quantitatively and qualitatively outperforms other techniques.

\end{abstract}


\section{Introduction}

Most physical tasks in the world, performed by humans or other animals, require being adept at multiple skills. For example, a lizard hunting prey may need to be proficient at climbing trees and running; a duck in migration needs to be able to both fly and swim; a human hurdler must be fast at running along bends and straightaways as well as jumping. 
If such animals were only capable of single motions, we would not expect them to be very successful.

We similarly should expect diverse adroitness from our robots.  Yet, recent successes in algorithms which can co-design robots over morphology and control have been typically catered to singlular task specifications.  For example, an algorithm may be able to co-design robots for forward running speed, energy efficient gaits, or climbing rough terrains, but not all three skills at once.  In order to computationally develop robots capable of composite tasks rather than single repetitious motions, we require algorithms that simultaneously optimize over collections of requisite skills. 

This work presents a method for multi-objective rigid robot co-design over both control and morphology (including discrete topology). Unlike much previous work on multi-objective co-design which only examined continuous parameters, we consider discrete topology as well as continuous control parameters. 
Because form informs function and vice versa, it is natural that different robot designs will be better at different tasks.  But, as in nature,
rarely will a single design be best at all tasks.  Thus, our goal is to extract robots with optimal trade-offs across different design objectives; \emph{i.e.} the Pareto set of robot designs for the tasks at hand.

Our method builds upon RoboGrammar \cite{zhao2020robogrammar}, a method that proposes a bio-inspired grammar for robot topological design and employs a learning-based morphological search over that design space. Robots are co-designed over topology and control; specifically, it employs a model predictive control (MPC) scheme.  
Complex grammars like that of RoboGrammar can be very expressive but typically yield large search spaces that are intractable to 
optimize over \emph{via} 
na\"ive methods; the problem only becomes more difficult when multiple objectives are considered.
The algorithm we present here is the only proposed method for solving this difficult and important multi-objective, topology/control co-design problem; thus making it novel in both problem scope and solution.

In this work, we contribute: \emph{1)} Multi-objective co-design algorithms for finding Pareto-optimal robot topologies and controllers over challenging objective trade-offs, \emph{2)} Demonstrations on combinations of terrestrial robot locomotion tasks, some with design restrictions, and \emph{3)} Comparisons of our proposed methods benchmarked against baselines, demonstrating the power and importance of our techniques.

\section{Related Work}

\subsection{Multi-objective Optimization}
Multi-objective search algorithms have been successfully deployed in a wide variety of engineering domains, including material design \cite{ashby2000multi}, automotive engineering \cite{liao2008multiobjective}, thermodynamics \cite{fettaka2013design}, and medicine \cite{nicolaou2013multi}, to name a few. Core to these applications is the development of search algorithms that can retrieve dense Pareto fronts that are close to the ground-truth, with high sample efficiency. Two popular categories of strategies exist. The first is evolutionary algorithms; see \cite{deb2001multi} for an introduction.  Popular methods in this space include NGSA-II \cite{deb2000fast}, NEAT \cite{stanley2002evolving}, and CMA-ES \cite{igel2007covariance}. These algorithms employ principled heuristics to efficiently trade off exploration of the design space with exploitation of the estimated Pareto front. Contrasted with evolutionary approaches are analytical methods such as \cite{eichfelder2009adaptive} or \cite{schulz2018interactive},  which combine probabilistic search with local, gradient-based optimization for increased efficiency. Particularly popular in this space are scalarization approaches (like \cite{eichfelder2009adaptive}), which perform continuous optimizations over sampled weight combinations. Our approach is similar to scalarization methods, but makes significant adaptations in order to account for large, grammar-based design spaces with stochastic objectives.

\subsection{Multi-Objective Control Optimization}
Multi-objective optimization has been recently applied to robot control problems through the combination of reinforcement learning methods and classical multi-objective optimization techniques. 
Methods such as \cite{gabor1998multi} and \cite{mannor2002steering} apply scalarizations to reduce these multi-objective searches to single objective problems that can be solved \emph{via} reinforcement learning algorithms.  Another category of methods discovers the entire set of control policies to approximate the true Pareto-optimal set, either through preference sampling \cite{6889738, li2019deep}, evolutionary algorithms \cite{xu2020prediction}, or universal control policy representation \cite{abels2018dynamic, NIPS2019_9605}. Compared to those methods, our work focuses on co-design of complex topology and control, and understanding their tight interplay.

\subsection{Robot Co-Design}

Robot co-design techniques 
include evolutionary approaches, analytical methods, and search-based approaches.

Evolutionary approaches apply evolutionary algorithms to iteratively improve robot form and behavior across generations. Early notable work in this space includes \cite{sims1994evolving} and \cite{hornby2003generative}, which evolved rigid robot morphologies and controllers to produce agile creatures.
It was recently demonstrated that such techniques could be applied to the robust space of neural network controllers  \cite{wang2019neural}. Such techniques have also been applied to soft robots \cite{cheney2013unshackling, cheney2014evolved, corucci2016evolving}, evolving robots over geometry, actuation, and open-loop control.  Evolutionary approaches lead to high diversity but poor convergence.

Analytical methods 
use 
model gradients in order to inform search. Analytical methods improve search efficiency, but can get stuck in local suboptima.
Rigid co-optimization over continuous morphological and control parameters for  walking robots has been extensively studied over the last decade \cite{wampler2009optimal, spielberg2017functional, ha2017joint, schaff2019jointly, geilinger2018skaterbots}. 
Similar continuous parameter searches have also been applied to soft robots
\cite{hu2019chainqueen, spielberg2019learning}.  

Search-based methods can both handle combinatorial topology search and still have promising efficiency. In these methods, discrete searches of joint and limb configurations are guided by heuristic functions. These heuristics can be provided \emph{a priori} \cite{ha2018computational} or, most similar to our work, iteratively learned \cite{zhao2020robogrammar} from evaluated designs.
\section{Preliminaries}

Our method builds upon RoboGrammar, a grammar-based search method for co-designing robot morphologies and controllers. Choosing a grammar as a search space enables one to search over discrete operations that define robot topology; this provides for much more expressive designs than afforded by purely continuous parameters. We refer the reader to \cite{zhao2020robogrammar} for details regarding the RoboGrammar method and briefly summarize its key components here.

\begin{figure}[t]
  \centering
  \includegraphics[width = 1.0\linewidth]{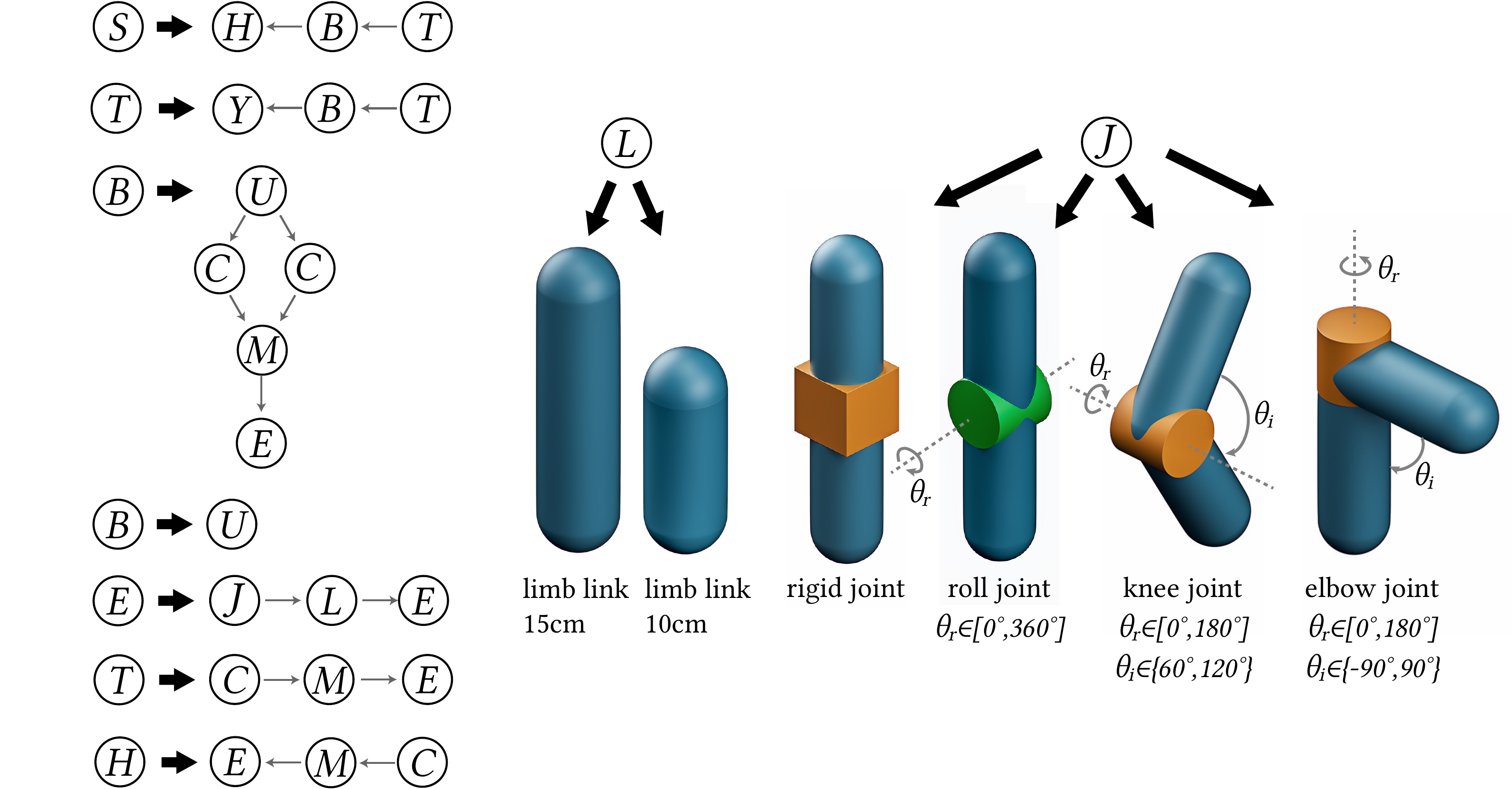}
  \caption{A graph-based grammar for terrestrial robot proposed by RoboGrammar.  \textbf{Left:} A subset of the grammar's rules.  \textbf{Right:} Physical components that can be generated by next-to-terminal symbols. The full grammar can be found in \cite{zhao2020robogrammar}.}
  \label{fig:grammar}
  \vspace{-15pt}
\end{figure}

In RoboGrammar, each robot topology design is represented as a directed acyclic graph, where each graph node is corresponding to a physically realizable component and each graph edge is corresponding to a physical link between those parts. RoboGrammar's morphological design space is then defined by a context-free graph grammar, and is constructed in order to promote terrestrially locomoting designs inspired by arthropods. Fig. \ref{fig:grammar} provides a snapshot of some of the most salient operations it describes. Each instance of the grammar has a graph representation where nodes can correspond to terminal or nonterminal symbols.  Starting from a nonterminal symbol ``S'', the grammar iteratively applies applicable rules to replace a nonterminal symbol in the current topology graph by a subgraph. A terminal design is produced once the graph consists of only terminal symbols. The grammar provides a constrained (\emph{i.e.} valid) but expressive design space. While we use this grammar for these listed advantages, we stress that the search algorithm presented in this paper is general and can be applied to any grammar-defined robot search space.

To search over its combinatorial design space, RoboGrammar proposes Graph Heuristic Search (GHS) algorithm, which is inspired by reinforcement learning. The algorithm alternates a heuristic-function-guided stochastic sampling of robot designs with a learning phase which improves that heuristic function.
The heuristic function's goal is to accurately predict the ``reward'' that a design will provide, including those of partial (non-terminal) designs generated during the design generation process. Despite the search space being combinatorial, the GHS prioritizes the branches that have the best chance of producing well-performing deigns; readers may draw parallels to A$^{*}$ search.
This heuristic is defined by a graph neural network architecture 
which naturally maps design's links to nodes and joints to edges.  The heuristic can operate on both complete and partial designs, enabling its use throughout the search.

In order to evaluate sampled designs and generate a control sequence, RoboGrammar employs MPC, 
specifically a stochastic MPPI scheme based off the POLO algorithm \cite{lowrey2018plan}. 
This stochasticity allows for exploration in the control planning stage and is resilient to poor local minima; however, this comes at the cost of noise in the evaluation procedure. Applying MPC twice to a morphology can lead to different controllers with very different 
rewards. This stochasticity requires care in the design generation phase, as it means the GHS must (judiciously) revisit designs to see if better controllers are feasible.

We note that although this co-design search may seem to be just two separate, sequential steps --- a morphological design search followed by control generation --- it is more accurate to describe it as a three-phase alternating co-optimization process when the heuristic is involved.  First, a heuristic is used to ``intelligently'' sample a design; 
second, the design is fed to MPC, and a motion is optimized; third, the reward calculated from the MPC-generated trajectory is used to improve the accuracy of the learned heuristic function. This three-phase process is visualized in Fig. \ref{fig:overview}. 

The remainder of this paper is structured as follows.  First, we introduce the scalarization approach to multi-objective optimization, and present two approaches to robot co-design which leverage this approach: a simple method based on solving a set of standalone subproblems, and a more sophisticated and efficient Multi-Objective Graph Heuristic Search (MOGHS) that uses GHS to iteratively expand the Pareto front along different sampled directions in objective space, while learning a heuristic function shared across these directions.  This shared heuristic is key to making our search efficient.
Then, we present experiments demonstrating the efficacy of our methods compared to baselines, 
and conclude with possible extensions to this work.

\section{Method}
\label{sec:method}

\newcommand{\weight}{\boldsymbol{\omega}}
   
Let $\pmb{\mathcal{D}}$ define the space of valid designs (morphology and control sequence).
We define a multi-objective function $\mathbf{F} : \pmb{\mathcal{D}} \rightarrow \mathbb{R}^{m}$, such that for $d \in \pmb{\mathcal{D}}$, $\mathbf{F}(d) = (f_1(d), f_2(d), ..., f_m(d))$. 
Our goal is to generate designs which are Pareto optimal in the objective space; in other words, designs which are \emph{non-dominated} by any other discovered design for any objective trade-off.  
In our maximization problem setting, a design $d$ is said to be Pareto optimal if $\nexists d' \in \pmb{\mathcal{D}} \text{ s.t. }: \forall i\,\,\,\, f_i(d') \geq f_i(d) ~ \wedge ~ \exists i \text{ s.t. } f_i(d') > f_i(d)$.    In layman's terms, a design is considered Pareto optimal if there does not exist another design that is strictly better than this design at all objectives.  We call the set of objective values of Pareto optimal designs the \emph{Pareto front}.  

\begin{wrapfigure}{R}{0.22\textwidth}
    \vspace{-12pt}
    \includegraphics[width=0.21\textwidth]{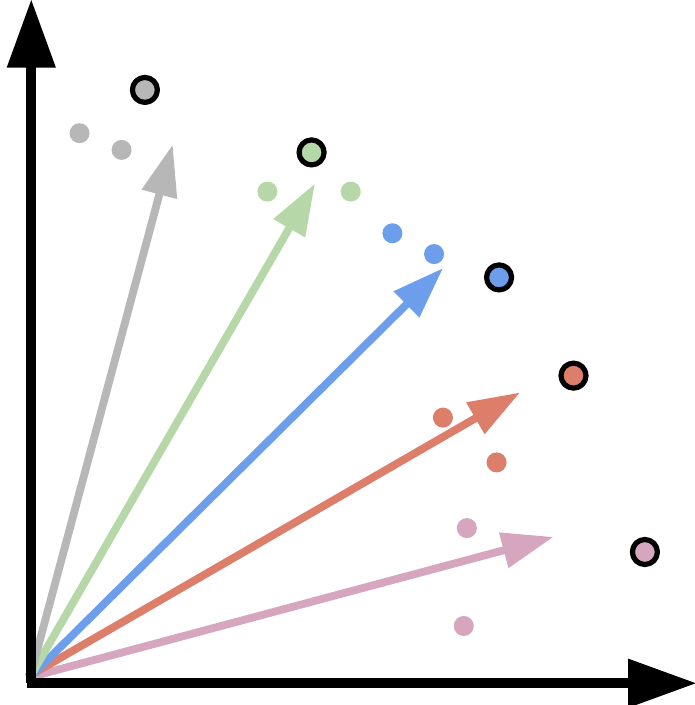}
    \caption{A cartoon depicting the scalarization method.  Weight pairs form rays that project radially outward from the origin.  Each circle represents a point that might be found during a single objective optimization using the weights defined by the ray of its color.  Circles with black borders are the optimal solutions to the corresponding weights, which form a convex Pareto approximation front.
    }
    \vspace{-10pt}
    \label{fig:scalarization}
\end{wrapfigure}

In practice, computing the exact Pareto set/front is intractable for most hard problems; thus, we seek an algorithm which can find a Pareto approximation set that is as ``good'' as possible.  We discuss quantitative metrics for evaluating the quality of a Pareto front in Sec. \ref{sec:results}.  However, we describe a qualitative way of determining the goodness of Pareto fronts here.  Consider weight vectors $\weight \in \mathbb{R}^m \emph{ s.t. } \forall i\,\,\,\, \omega_i \geq 0 \text{ and } \|\weight\|_p = 1$ for some norm $p$. This can be thought of as the space of rays that sweep out radially in the first orthant.  We wish to find a Pareto set such that, for every valid $\weight$, there exists a point $d$ in the Pareto set for which $\weight \cdot \mathbf{F}(d)$ is large. 
In other words, for each (scalarization direction) ray $\weight$  we want to find points whose objectives are far away from the origin along that ray.

Given this,
an obvious strategy for multi-objective optimization arises, termed \emph{scalarization} methods. For a large collection of weight vectors $\{\weight^i\}_{i=1}^n$, a Pareto approximation set can be extracted by solving the set of optimization subproblems, where subproblem $\mathcal{P}_i$ is $\arg\max_{d \in \pmb{\mathcal{D}}} \weight^i \cdot \mathbf{F}(d)$ (Fig. \ref{fig:scalarization}). This approach leads to two challenges.
The first challenge is to find (as close to) the global maximum of each optimization subproblem.  The second challenge is to have an efficient optimization scheme such that a dense set of weight vectors can be optimized.

As a first attempt at solving this problem, we propose the following (na\"ive) ``discrete weights'' strategy.  Given a budget $n$, sample a uniformly spaced set $\{\weight^i\}_{i=1}^n$ \emph{a priori}.  Then, for each weight vector $\weight^i$, solve the $n$ corresponding $\mathcal{P}_i$ independently, using the approach presented in \cite{zhao2020robogrammar} as a black box with reward  $\weight^i \cdot \mathbf{F}$.
Such an approach can unfortunately have poor sample efficiency, especially in high-dimensional objective spaces, and treats problems with shared structure as decoupled. Thus, here we propose an alternative algorithm that is more effective at extracting good Pareto fronts. 

\begin{figure}[t]
  \centering
  \includegraphics[width = 1.0\linewidth]{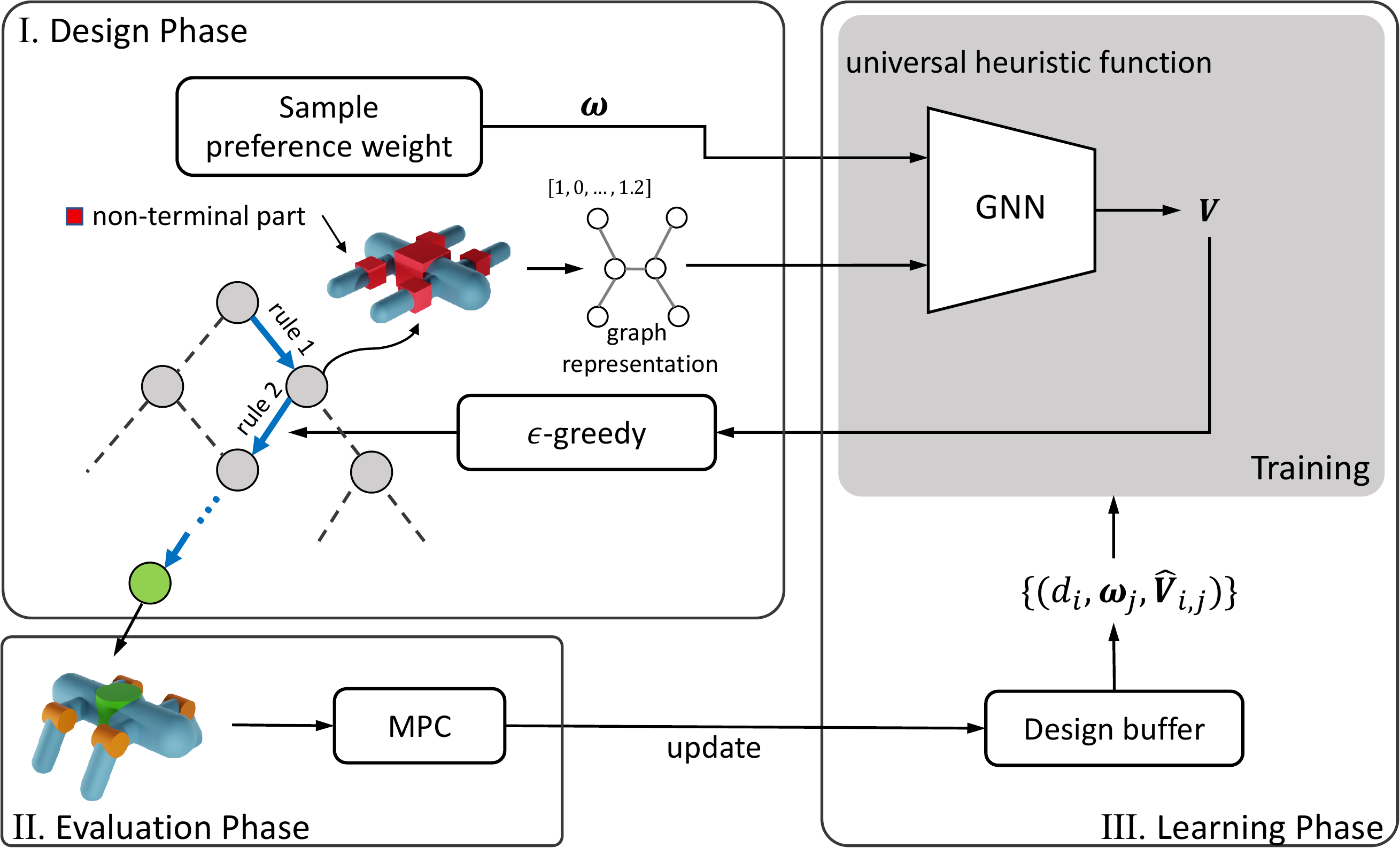}
  \caption{Overview of the Multi-Objective Graph Heuristic Search (MOGHS). In each episode, the algorithm conducts three phases (similar to GHS). \textit{Design Phase}: A robot design is selected using a learned universal graph heuristic function along with a randomly picked preference weight $\weight$. \textit{Evaluation Phase}: The selected robot design is evaluated by MPC for each objective. \textit{Learning Phase}: All the designs seen so far are leveraged to improve the heuristic.}
  \label{fig:overview}
  \vspace{-12pt}
\end{figure}

\subsection{Multi-objective Graph Heuristic Search} 
The discrete weights strategy has two shortcomings.  The first  is its decision to fix weights \emph{a priori}.  This makes it harder to find Pareto optimal points that do not lie along the sampled scalarization directions. The second shortcoming is the decision to treat each subproblem independently, despite the clear shared structure between the subproblems.

We thus improve the approach presented in RoboGrammar to efficiently search in multi-objective spaces. Our multi-objective optimization strategy makes the following two core changes. First, we sample weight vectors uniformly randomly at each episode of the algorithm both to search over the entire space of weight vectors in a \emph{single} invocation of search algorithm, as well as to have a \emph{dense} set of scalarization directions. 
Our second core innovation is to modify our learning model to be \emph{multi-objective} and \emph{universal} (named after its similarity in structure to universal control evaluators \cite{abels2018dynamic}), which we describe shortly. This improves search efficiency because the representations learned by the heuristic function can be shared by all weight combinations.

The process of MOGHS is visualized in Fig. \ref{fig:overview}, and the full algorithm can be found in Algorithm \ref{alg:moghs}.  Here, the original GHS is presented in black and teal as in Zhao et al. \cite{zhao2020robogrammar}, with our new modifications highlighted in red. We now describe the major changes to the three phases of RoboGrammar's GHS to formulate our MOGHS.

\subsubsection{Design Phase}
  In GHS, a designs is sampled in each episode according to a ``double $\epsilon$-greedy'' approach in order to balance exploration and exploitation.  First, $K$ designs are sampled by  $\epsilon$-greedily choosing the next rule to apply at each step of the generation process, where the greedy choice is chosen by the one with the best heuristic function value.  The final design can then be chosen greedily or rejected for a random one at an $\epsilon$ rate.  In MOGHS, we instead first sample an $\weight$ vector uniformly at random, a search direction for this episode of the design search.  Using this, the greedy selection scheme is generalized to multiple objectives by the linear scalarization $\weight \cdot \mathbf{V}(d, \weight)$.  

  
  Readers may note that the new greedy selection criterion requires two changes to the structure of the heuristic function.  First, the heuristic must accept a weight vector $\weight$ in addition to a design $d$ as input, thus making it a \emph{universal} predictor among all possible $\weight$.  Second, it outputs a vector of the predicted value of each objective function when evaluated by MPC, thus making it \emph{multi-objective}.  Thus, we design such a heuristic function $\mathbf{V} \colon \pmb{\mathcal{D}} \times \mathbf{R}^m \rightarrow \mathbf{R}^m$.

\subsubsection{Evaluation Phase}
A candidate design is evaluated $m$ times, in $m$ separate invocations of MPC. The $i^{th}$ invocation is run optimizing reward (objective) $f_i$. This returns $m$ different optimized control sequences for the sampled design\footnote{If $f_i$ does not depend on the robot's motion (\emph{e.g.} the design complexity objective in section \ref{sec:exp}),
this process can be skipped for that objective.}.

\subsubsection{Learning Phase}
Given sampled batch of designs $S_b = \{d_i\}$ and weights $W = \{\mathbf{\omega}^j\}$, we first compute the target heuristic value $\hat{\mathbf{V}}_{ij}$ for each design-weight combination. Then we train our universal heuristic function by using Adam \cite{DBLP:journals/corr/KingmaB14} to minimize the following loss function:
\begin{equation}
    \sum_{i,j}\|\mathbf{V}(d_i, \mathbf{\omega}^j) - \hat{\mathbf{V}}_{ij}\|_2^2
\end{equation}
The target heuristic value $\hat{\mathbf{V}}$ is computed by maintaining a Pareto front of rewards for each partial design to keep track of the optimal terminal designs can be induced from it. 

\subsection{Universal Multi-Objective Heuristic Function}
Our heuristic function is a graph neural network which maps robot morphologies to output vectors of predicted rewards. The morphology is represented as a graph, where each node corresponds to a link of the robot (and whose corresponding feature vectors describe their geometric and inertial properties), and each edge corresponds to a joint, thus encoding the topology. The architecture used is based on the differentiable pooling architecture presented by \cite{ying2018hierarchical}.  

We wish for this network to operate on both a robot morphology as well as a weight vector as input; however, this architecture only handles graphs.  In order to learn latent representations in the network which include the effects of the objective weights, we include $\weight$ as additional features to each node in the graph.  
Finally, we modify the final linear operator of the network to output $m$ channels instead of one, thus providing a multi-objective output space of the heuristic.

\subsection{Other Improvements}

We include the following further modifications which improve sample efficiency:

\subsubsection{DAG-Based Target Updates}
\label{sec:dag}
In \cite{zhao2020robogrammar}, when a design is evaluated, its value is propagated up the derivation path to update the target value of the partial designs that generated it.
Realizing the design rules actually form a directed acyclic graph (DAG) --- each partial or complete design can have many rule sequences that generate it --- we now perform the upward propagation procedure even on non-evaluated designs, which provides an opportunity to merge previously visited partial designs up newly discovered branches of the DAG.
This simple refinement vastly improves sample efficiency by improving the accuracy of the computed target values.  

\subsubsection{Invalid Design Marking}
Although the grammar is designed to avoid invalid designs, they still can occur.  Once we know a design is invalid, however, we need not visit it again.  We mark all invalid designs as such, and remove them from the pool of candidates to generate.  If all designs that could be generated by a partial design are invalid, we further mark that partial design as invalid in an upward propagation scheme similar to the DAG-based update.

\subsection{Implementation}

Despite MOGHS requiring many samples over design and control, the algorithm provides many opportunities for parallelization over CPU cores, thus keeping it practical.
First, MOGHS samples many designs in parallel, which can be parallelized over many CPU cores.  Second, the main bottleneck of MOGHS is the evaluation phase; fortunately, our MPC algorithm is sampling-based; this sampling procedure can also be parallelized over CPU cores to improve efficiency.  Finally, the learning procedure can be accelerated by batching $M N_{W}$ samples in parallel. Running 2000 episodes of MOGHS takes approximately 20 hours on a 64-core Google Cloud machine, and the breakdown for each phase is around 3 hours (using 16 cores) for the design phase, 11 hours (using 64 cores) for the MPC evaluation phase, and 6 hours (using 1 core) for the learning phase.

\begin{algorithm}
\small
\begin{algorithmic}
\STATE \textbf{Inputs:} Number of iterations $N$, number of candidate designs $K$, Adam optimization steps opt\_iter and batch size $M$, number of sampled weights $N_w$.
\STATE \textbf{Output:} A set of Pareto-Optimal designs $P$.

 \STATE Initialize the universal graph neural network {\color{red!80}{$\mathbf{V}_\theta(d, \weight)$}}.
 
 \STATE Initialize the Pareto-Optimal design set $P \leftarrow \{\}$.

 \FOR{episode $j \leftarrow$ $1$ \TO $N$}
 
    \STATE \color{teal}{$\rhd$ \textbf{Design Phase:} Generate a candidate design}
    
    \STATE {\color{red!80}{Sample a preference weight $\weight$.}}
    
    \STATE{\color{black}{$C \leftarrow \{\}$} \hfill \color{teal}{$\rhd$ Initialize possible design candidates}}
    
    \algorithmiccomment{Sample $K$ designs by $\epsilon$-greedy approach}
    
    \FOR{$k$ $\leftarrow$ 1 \TO $K$} 
        \STATE{$d \leftarrow $ initial design graph}
        
        \WHILE{$d$ has non-terminals} 
            \STATE With probability $\epsilon$ select a random rule $a$, otherwise select {\color{red!80}{$a = \arg\max_a \weight \cdot \mathbf{V}_{\theta}(d', \weight)$}}, where $d'$ is the robot design after applying rule $a$ on design $d$.
            
            \STATE $d \leftarrow d'$
        \ENDWHILE
        \STATE{Add possible candidate $d$ to $C$.}
    \ENDFOR
    
    \algorithmiccomment{Choose one to be the candidate}
    \STATE With probability $\epsilon$ select a random sampled design $d$ from $C$, otherwise select {\color{red!80}{$d = \arg\max_{d \in C} \weight \cdot \mathbf{V}_{\theta}(d, \weight)$}}.
    
    \algorithmiccomment{\textbf{Evaluation Phase:} Compute the rewards for the design}

    \STATE {\color{red!80}{Run MPC in parallel for each task to evaluate the rewards vector $\vec{r}$ of design $d$}.}
    
    \algorithmiccomment{Update the design Pareto set}
    \STATE Update $P$ by $d$ and $\vec{r}$.

    \algorithmiccomment{\textbf{Learning Phase:} train heuristic value function $\mathbf{V}_{\theta}$}
    \FOR{$i$ $\leftarrow$ 1 \TO opt\_iter} 
        \STATE Sample a minibatch $S_b$ of seen designs (partial or complete) of size $M$.
        
        \STATE {\color{red!80}{Sample $N_w$ preference weights $W = \{\weight^j\}$.}}
        
        \STATE {\color{red!80}{Compute target values for each $s \in S_b$ and $\weight \in W$,}} 
        \vspace{-0.5em}
        \STATE {\color{red!80}{$$\hat{\mathbf{V}}(s, \weight) =  \underset{d \in \text{descendant}(s)}{\arg\max} \weight \cdot \vec{r}(d) $$}} 
        \vspace{-0.5em}
        \STATE Update $\mathbf{V}_\theta(s, \weight)$ one step by Adam with the loss:
        \vspace{-0.5em}
        \STATE $$\sum_{s \in S_b, \weight \in W} \|\mathbf{V}_\theta(s, \weight) - \hat{\mathbf{V}}(s, \weight)\|^2$$
        \vspace{-0.5em}
    \ENDFOR
 \ENDFOR

 \caption{Multi-Objective Graph Heuristic Search}
 \label{alg:moghs}
 \end{algorithmic}
\end{algorithm}
\setlength{\textfloatsep}{1em}
\section{Experiments}
\label{sec:exp}
We compare the Pareto fronts discovered by three algorithms: \emph{1)} a random baseline, in which designs are sampled by randomly selecting rules until a terminal design is generated 
, \emph{2)} The discrete weights method proposed in \ref{sec:method}, which is a discrete version of our MOGHS algorithm, and \emph{3)} our MOGHS algorithm. We use the same total MPC evaluation budget (\emph{i.e.} number of evaluated designs) for all three algorithms.  We demonstrate our algorithm on six combinations of seven tasks, and compare each solution set qualitatively and quantitatively; please see the video for demonstrations of robots along the discovered Pareto fronts.

\emph{Flat Terrain Locomotion:}
In this task, the robot is rewarded for the forward running speed, and we assign additional reward to encourage stability in the forward direction.

\emph{Low Power Flat Terrain Locomotion:}
The same as the Flat Terrain task, except the maximum impulse of the motors is set to $20\%$ of that normally available. This task highlights  locomotion in scenarios when power must be conserved.

\emph{Wall Terrain Locomotion:}
Also the same as the Flat Terrain task, however, slalom-like walls are added to the terrain. Successful robots must run forward with the ability to move somewhat laterally to navigate terrain.

\emph{Jumping:}
In this task, the robot must jump as high as possible. The reward is set proportional to the height of the lowest part of the robot. As before, an additional reward is added to discourage the robot from falling over.

\emph{Spin-In-Place:}
This task  tests the agility of the robot around the vertical axis. The reward is set proportional to the angular velocity of the robot around the vertical axis.

\emph{Design Complexity:}
The first of two tasks that is purely design-dependent (does not involve control), the reward is set inversely proportional to the number of links in the robot.  

\emph{Robot Height:}
The second pure design task, the reward is set proportional to the height of the robot, with a penalty for changes in pitch, promoting tall, upright robots.

\subsection{Experimental Setup}

We run each experiment for $2000$ episodes.  For each task combination, we run each algorithm three times.  In comparing metrics in Table \ref{tab:hv}, we compute the metric by averaging over all runs for that algorithm.  For metrics that require a reference set, we take the union of all sampled designs of all runs of all algorithms, and compute its Pareto front.  Hyperparameters used for the MOGHS algorithm are the same as GHS \cite{zhao2020robogrammar}, with the preference weight minibatch size $N_{W}$ set to $10$.  For the discrete weights algorithm, we sample $11$ uniform weights in the two-objective cases (we did not consider this baseline in the three-objective case).

\begin{figure*}[h]
    \centering
    \includegraphics[width=0.95\textwidth]{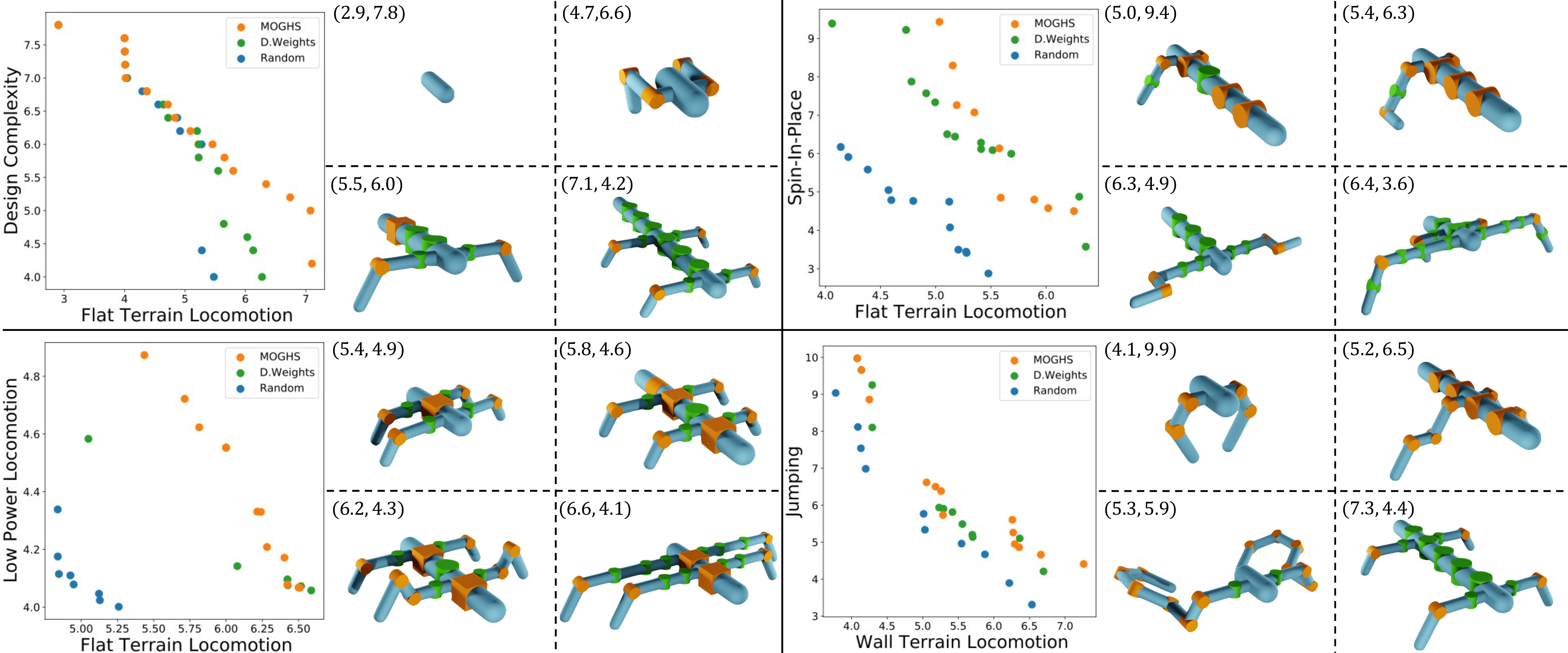}
    \caption{Pareto front comparison of four of our two-objective experiments, and example designs from the Pareto front.  MOGHS produces more diverse and better performing results than discrete weights or the random baseline.}
    \label{fig:pareto_designs}
    \vspace{-1em}
\end{figure*}

\subsection{Results}
\label{sec:results}

We numerically evaluate the optimized Pareto fronts on three metrics, commonly used in the multi-objective optimization literature \cite{chen2020evaluate}: Hypervolume, Generational Distance, and Inverse Generational Distance.  We present some Pareto fronts in Fig. \ref{fig:pareto_designs}, along with some selected designs.  We encourage the reader to watch the video for more renderings of optimized Pareto fronts, and animations of the designs that populate them.

\,\emph{a) Hypervolume} \cite{zitzler1999multiobjective}:\,
The hypervolume metric (HV) measures the hypervolume of the polytope defined by the space enclosed by the hyperplanes created by the axes along the first orthant and the hull of the sampled points.  
As is visually evident,  a larger HV is better.

\,\emph{b) Generational Distance} \cite{van1999multiobjective}:\,
The generational distance (GD) of a Pareto front $P$ is defined as:
\begin{align}
    GD(P) = \frac{1}{|P|} ( \sum_{\mathbf{p} \in P} \min_{\mathbf{r} \in R} d(\mathbf{p}, \mathbf{r})^p )^{\frac{1}{p}}
\end{align}
where $p$ is a norm (we choose $1$), $d$ is the Euclidean distance function, each $\mathbf{p}^i$ is a point in $P$, and the set $R$ is a reference set generated by combining results from all optimization experiments.  A smaller GD is better.

\,\emph{c) Inverse Generational Distance} \cite{10.1007/978-3-540-24694-7_71}:\,
The inverse generational distance (IGD) is defined as:
\begin{align}
IGD(P) = \frac{1}{|R|} \sum_{\mathbf{r} \in R} \min_{\mathbf{p} \in P} d(\mathbf{r}, \mathbf{p})    
\end{align}
Again, a smaller IGD is better.  Roughly speaking, GD measures the distance of all points on the captured Pareto front to the \emph{best} known Pareto front, while IGD measures the distance of all points on the \emph{best} known Pareto front to the captured Pareto front.

Numerical results are presented in Table \ref{tab:hv}.  As can be seen in MOGHS dominates discrete weights in all task combinations across all metrics, often by significant margins.  Discrete weights, in turn tends to beat the random baseline, but not as consistently.  The consistency and quality of results returned by MOGHS emphasizes the importance of this method.
Qualitatively, MOGHS's Pareto fronts, as seen in Fig. \ref{fig:pareto_designs} tend to yield better performing objective trade-offs than other methods, while maintaining dense coverage.

The morphologies and motions of the designs found on the Pareto fronts of each problem are physically principled, but still interesting and exciting.  We consider four of the two-objective trade-offs here.  We leave the Design-Height problem, which serves as a benchmarking example, and the complex three-objective Flat-Jump-Spin task, which is better visualized animated, for the video.  For example, in the Flat-Design task, a wide spectrum of robots from a single link (simplest design but no motion) to long, complex, fast, walkers are recovered.  Along the way, various slower walkers with fewer links are found along the front, with increasingly dynamic motions.  The Flat-LowPower task measures robots in various stages of power consumption.  In the low power configuration, the robot is unable to balance if the legs are too wide, due to the increase gravitational torque on the torso.  This leads to poor forward locomotion.  However, longer legs lead to faster strides for the normal power state robots.  The trade-off provides a spectrum of robots of varying width and compactness, trading off the importance of being effective at forward locomotion in the two power states.  The Flat-Spin trade-off produces a very exciting and surprising result, as faster spinners trade off forward locomotion skill for an ability to spin in a top-like fashion.  The fastest spinners possess motions that resemble breakdancing.  The Wall-Jump robots meanwhile transition from maneuverable walkers to increasingly frog-like morphology with increased capacity to hop.

\begin{table}[t]
\caption{  \begin{singlespace}A comparison of the three numerical metrics among all three algorithms presented.  For each problem, metrics are presented in the following order: HV, GD, IGD.  Bolded numbers mean that column's algorithm performed best for that algorithm/problem combination. MOGHS outperforms other methods in all metrics across all problems \end{singlespace}}

\label{tab:hv}
\begin{center}
\begin{small}
\vskip -0.2 in
\scalebox{0.95}{
\begin{tabular}{ccccc}
\toprule
\sc{Problem}  & & \sc{Moghs} & \sc{D. Weights} & \sc{Random}\\
\midrule
\multirow{3}{*}{Design-Height}& \scriptsize{Hv} &\textbf{64.70} & 52.43 & 51.60 \\
 & \scriptsize{GD} & \textbf{0.04} & 0.45 & 0.29 \\
 & \scriptsize{IGD} & \textbf{0.28} & 1.02 & 0.96 \\
\hline
\multirow{3}{*}{Flat-Design} & \scriptsize{Hv} &\textbf{46.30} & 42.25 & 39.11 \\
 & \scriptsize{GD} &\textbf{0.10} & 0.29 & 0.31 \\
 & \scriptsize{IGD} &\textbf{0.19} & 0.41 & 0.56 \\
\hline
\multirow{3}{*}{Flat-Spin} & \scriptsize{Hv}&\textbf{49.94} & 46.90 & 29.88 \\
 &\scriptsize{GD}& \textbf{0.34} & 0.40 & 1.35 \\
 &\scriptsize{IGD}& \textbf{0.37} & 0.50 & 1.79 \\
\hline
\multirow{3}{*}{Flat-LowPower} & \scriptsize{Hv}&\textbf{29.94} & 28.03 & 22.25 \\
 &\scriptsize{GD}& \textbf{0.04} & 0.26 & 0.92 \\
 &\scriptsize{IGD}& \textbf{0.08} & 0.28 & 1.07 \\
\hline
\multirow{3}{*}{Wall-Jump}  &\scriptsize{Hv}& \textbf{57.95} & 54.09 & 44.11 \\
&\scriptsize{GD} & \textbf{0.37} & 0.73 & 1.23 \\
 &\scriptsize{IGD }& \textbf{0.36} & 0.64  & 1.15 \\
\hline
\multirow{3}{*}{Flat-Jump-Spin} &\scriptsize{Hv} & \textbf{307.68} & - & 166.71 \\
 &\scriptsize{GD}& \textbf{0.22}  & -  & 1.09 \\
 &\scriptsize{IGD}&  \textbf{0.38} & - &  1.69 \\
\bottomrule
\end{tabular}
}
\end{small}
\end{center}
\vspace{-1em}
\end{table}
 
\section{Conclusion}

We have demonstrated methods for co-designing robots over morphology and control over multiple objectives.  Our multi-objective graph heuristic search is first of its kind, and extracts far superior Pareto fronts with higher efficiency than more na\"ive methods.  The tasks demonstrated, including running, jumping, spinning, and obstacle navigation  have direct practical value in real-world terrestrial agile robots.

There remain important avenues for future research.  First, all examples demonstrated in this paper were tested in simulation.  A study fabricating these designs and demonstrating their physical accuracy would be valuable.  Second, all examples presented in this paper were for two or three objective trade-offs.  It would be interesting to see if this algorithm would scale to higher objective spaces.
Finally, although we have demonstrated our algorithm for robotics, most of our method should be general to any grammar-defined domain, excepting the evaluation (control) procedure and choice of heuristic architecture.  Extensions of our algorithm to other domains would 
demonstrate further value of our approach.
\section*{Acknowledgements}
We thank the anonymous reviewers for their helpful comments in revising the paper. This work is supported by Intelligence Advanced Research Projects Agency (grant No. 2019-19020100001), and Defense Advanced Research Projects Agency (grant No. FA8750-20-C-0075).

\bibliographystyle{IEEEtran}
\bibliography{main}

\end{document}